%% file: main.tex
\documentclass[letterpaper, 10 pt, journal]{ieeeconf}

\IEEEoverridecommandlockouts                              

\usepackage{graphicx}
\usepackage{amssymb}  
\usepackage{hyperref,tcolorbox}
\usepackage{subcaption}
\usepackage{enumerate}
\usepackage{cite}
\usepackage{amsmath,bm}
\usepackage[algo2e]{algorithm2e} 
\usepackage{soul} 
\usepackage[normalem]{ulem}
\usepackage{siunitx}

\newif\ifshortversion
\shortversiontrue

\newif\ifresubmissionred
\resubmissionredfalse

\newif\ifmodification
\modificationfalse

\newcommand\norm[1]{\left\lVert#1\right\rVert}
\newcommand{\stkout}[1]{\ifmmode\text{\sout{\ensuremath{#1}}}\else\sout{#1}\fi}

\usepackage{scrextend}

\usepackage{comment}
\usepackage{titlesec}
\let\labelindent\relax
\usepackage{enumitem}
\usepackage{cancel}

\titlespacing*{\section}
{0pt}{2.0ex plus .1ex minus .5ex}{1.5ex plus .1ex}

\input lamacro

\title{\LARGE \bf
Fast and Accurate Relative Motion Tracking for Dual Industrial Robots
}

\def\dq{\delta \bm{q}}
\def\qdot{\bm{\dot q}}

\def\nustar{\nu^*}

\def\ez{e_z}

\def\ldot{\dot\lambda}
\def\lddot{\ddot\lambda}

\def\sq{^2}

\author{Honglu He$^\dagger$\thanks{$^\dagger$  Rensselaer Polytechnic Institute, Troy, NY. Emails: {\tt heh6@rpi.edu, luc6@rpi.edu, saundg@rpi.edu,paters@rpi.edu, juliua2@rpi.edu, wenj@rpi.edu}},
        Chen-lung Lu$^\dagger$,
Glenn Saunders$^\dagger$, 
Pinghai Yang$^*$\thanks{$^*$GE Aereospace, Niskayuna, NY. Emails: {\tt Pinghai.Yang@ge.com, schoonov@ge.com, ajdelszt@ge.com}}, 
Jeffrey Schoonover$^*$, Leo Ajdelsztajn$^*$, \\
John Wason$^\ddagger$\thanks{$^\ddagger$Wason Technology, Tuxedo, NY. Email: {\tt wason@wasontech.com}},
Santiago Paternain$^\dagger$,
Agung Julius$^\dagger$, John T.~Wen$^\dagger$ 
}

\begin{document}
\maketitle
\thispagestyle{empty}
\pagestyle{empty}

\begin{abstract}

%

Industrial robotic applications such as spraying, welding, and additive manufacturing frequently require fast, accurate, and uniform motion along a 3D spatial curve. To increase process throughput, some manufacturers propose a dual-robot setup to overcome the speed limitation of a single robot.  Industrial robot motion is programmed through waypoints connected by motion primitives (Cartesian linear and circular paths and linear joint paths at constant Cartesian speed).  The actual robot motion is affected by the blending between these motion primitives and the pose of the robot (an outstretched/near-singularity pose tends to have larger path-tracking errors).  Choosing the waypoints and the speed along each motion segment to achieve the performance requirement is challenging.  At present, there is no automated solution, and laborious manual tuning by robot experts is needed to approach the desired performance. 
In this paper, we present a systematic three-step approach to designing and programming a dual-robot system to optimize system performance.  The first step is to select the relative placement between the two robots based on the specified relative motion path.  The second step is to select the relative waypoints and the motion primitives. The final step is to update the waypoints iteratively based on the actual 
\ifresubmissionred{\color{red}\fi
measured
\ifresubmissionred}\fi
relative motion. Waypoint iteration is first executed in simulation and then completed using the actual robots. For performance assessment, we use the mean path speed subject to the relative position and orientation constraints and the path speed uniformity constraint.   
\ifresubmissionred{\color{red}\fi
We have demonstrated the effectiveness of this method on two systems, a physical testbed of two ABB robots and a simulation testbed of two FANUC robots, for two challenging test curves.
\ifresubmissionred}\fi

\end{abstract}

\section{INTRODUCTION}
\label{sec:intro}
\begin{figure}[ht]
     \centering
     \begin{subfigure}[b]{0.21\textwidth}
         \centering          
        \includegraphics[width=\textwidth]{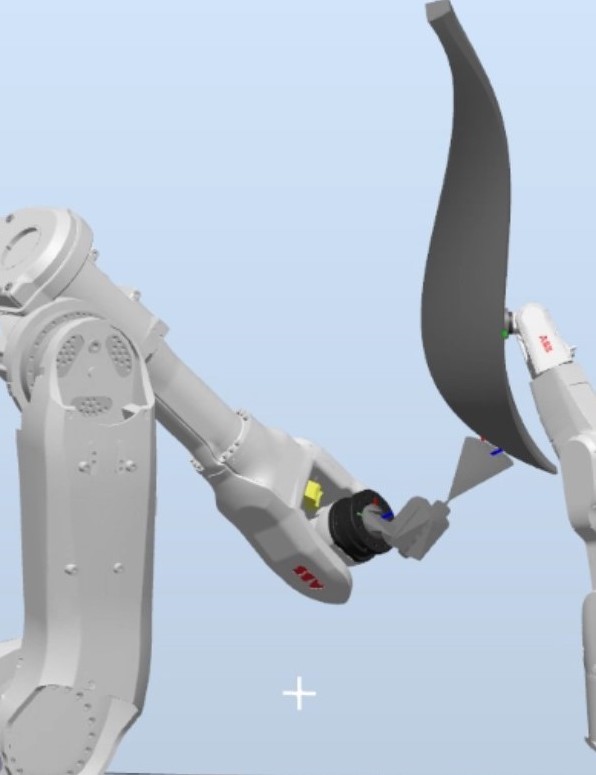}
         \caption{\small{Spraying in Simulation}}
         \label{fig:curve2_sim_abb}
     \end{subfigure}
    \begin{subfigure}[b]{0.229\textwidth}
         \centering                      \includegraphics[width=\textwidth]{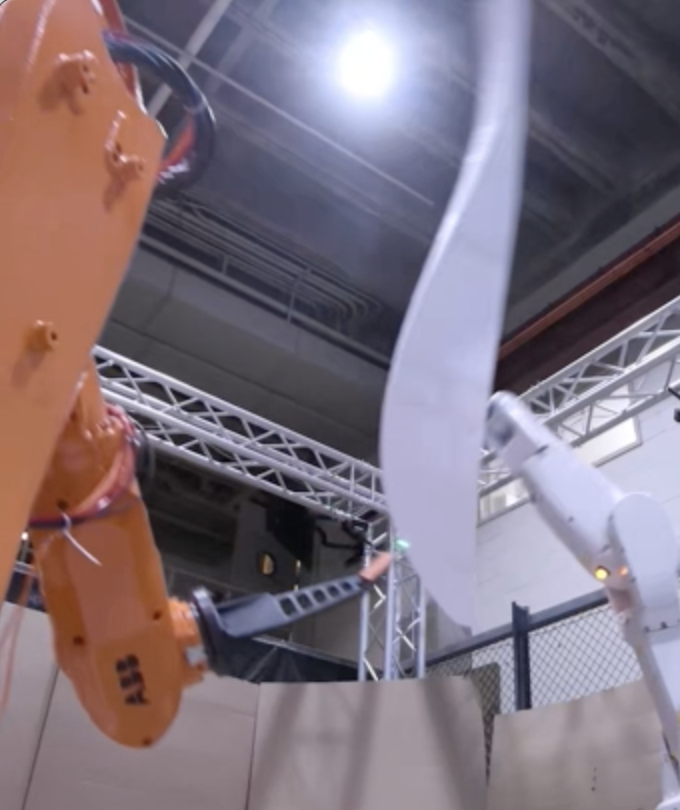}
         \caption{\small{Spraying Mock-up}}
         \label{fig:curve2_phy_abb}
     \end{subfigure}
        \caption{\small{Dual-arm spraying mock-up in simulation and physical testbed.}    
    }      
    \label{fig:curve2_abb}
\end{figure}

Industrial robots are increasingly deployed in applications such as spray coating \cite{spray_coating}, arc welding \cite{arc_welding}, deep rolling \cite{deep_rolling}, surface grinding \cite{surface_grinding}, cold spraying \cite{cold_spray}, etc., where the tool center point (TCP) frame attached to the end effector needs to track complex geometric paths in both position and orientation. In most applications, the task performance is characterized by the motion speed (how long it takes to complete the task), motion uniformity (how much the speed varies along the path), and motion accuracy (the maximum position and orientation tracking errors along the path). 
\ifresubmissionred{\color{red}\fi
To achieve the desired performance (high traversal speed subject to tracking accuracy and motion uniformity), some manufacturers are exploring dual-arm coordinated motion for relative spatial curve tracking, such as cold spray coating of the metal leading edge of a turbine fan blade \cite{GEcoldspray}, where one arm holds the spray gun and the second arm holds the blade.  A mock-up of the same operation in simulation and in our lab is shown in Fig.~\ref{fig:curve2_abb}.
\ifresubmissionred}\fi

\ifresubmissionred{\color{red}\fi
Dual-arm manipulation has long been studied in the robotics literature, but mostly for bimanual manipulation where the two robots and the load form a closed kinematic chain \cite{NN_bimanual}.  Early work considers torque level control to manipulate the load and the contact force \cite{hayati1986hybrid}. Most industrial robots do not allow direct torque control; instead, they provide a position or velocity setpoint streaming option (e.g., External Guided Motion by ABB, Streaming Motion by FANUC, MotoPlus by Yaskawa Motoman, etc.).  This feature has been used to implement resolved rate control, including early work on PUMA robots \cite{zheng1989integrating} and more recent work on bimanual manipulation using quadratic programming in \cite{four-criterion,multicriterion}.
Dual-arm motion planning algorithm for obstacle avoidance has also been extensively studied, e.g., \cite{shi2022obstacle,RRTSmart-AD}, again in the context of a closed kinematic chain.  

This paper differs from past dual-arm work in that our focus is to optimize the relative motion between two robots to achieve fast and uniform spray deposition. Further, we only use standard motion primitives connecting specified waypoints instead of setpoint streaming.  All industrial robot vendors offer these motion primitives, while motion streaming requires additional packages or may not be available. Motion streaming has low sampling rates and incurs high latency; as a result, it has much worse trajectory tracking performance than the built-in robot controller used in the motion primitives. Single-arm trajectory optimization for path following has long been studied \cite{shin1985minimum}. There has been recent work on dual-arm minimum time path tracking using convex programming, but there is no speed uniformity requirement \cite{an2023time}. Dual-arm welding has addressed redundancy resolution and speed uniformity \cite{dual_welding} but there is no motion optimization.

\ifresubmissionred}\fi

\ifresubmissionred{\color{red}\fi
Our approach is to first formulate the exact path tracking problem as an optimization, maximizing the path speed subject to the relative robot path tracking specification and the joint velocity and acceleration constraints of each robot.  The optimization involves finding the configuration parameters (relative pose between the robots and initial robot joint angles) and resolving the kinematic redundancy subject to the relative path tracking constraint. 
We propose a solution strategy for the exact tracking problem by using differential evolution for the global optimization of the configuration parameters. The objective function for this optimization is the maximum uniform path speed subject to the exact path tracking constraint.  The maximum path speed is calculated by combining the Jacobian-based redundancy resolution with the joint velocity and acceleration constraints expressed as a path speed constraint.
%
The solution of the exact tracking problem is not directly implementable on industrial robots as the commanded motion is not necessarily the actual motion due to the performance of the proprietary robot controllers.  We, therefore, pose a relaxed version of the optimization problem to allow tolerances of path speed variation and relative path deviation. 
Typical industrial robot controllers only allow a small set of motion primitives, Cartesian linear and circular paths and joint paths ({\tt moveL, moveC, moveJ}) between waypoints under a specified commanded speed.  We solve the relaxed tracking problem by approximating the solution of the exact tracking problem using the available motion primitives and finding the highest path speed that satisfies all constraints.
%
This problem is currently tackled in the industry through iterative manual adjustment.  The contribution of this paper is to formulate the motion optimization control problem analytically, develop a solution strategy that is readily implementable on industrial robots, and demonstrate the feasibility both in simulation and on a physical testbed.

The rest of the paper is organized as follows: Section~\ref{sec:problem_formulation} presents the problem formulation based on the performance requirements.  Section~\ref{sec:approach} presents the solution approach.  Section~\ref{sec:results} discusses the implementation details and results.

\ifresubmissionred}\fi

\setlist[itemize]{leftmargin=*}

\def\armsup#1{^{(#1)}}
\def\q#1{\bm{q}^{(#1)}}
\def\qdot#1{\bm{\dot q}^{(#1)}}
\def\p#1{\bm{p}\armsup{#1}}
\def\R#1{\bm{R}\armsup{#1}}
\def\J#1{J\armsup{#1}}

\section{Problem Formulation}
\label{sec:problem_formulation}

\ifresubmissionred{\color{red}\fi

\noindent \textbf{Notation:}
\begin{itemize}
    
\item $\bm{p}\in \mathbb{R}^3, \bm{\bm{\beta}} \in \mathbb{R}^3$: Cartesian position and the angle-product representation of orientation.
\item $\bm{q}, \bm{\dot q} \in \mathbb{R}^6$: robot joint position and angular velocity. 
\item $\bm{\bm{T}_b}: 4 \times 4$ transformation matrix of robot base pose.
\item $\bm{R}: 3 \times 3$ rotation matrix.
\item $\bm{p}^*\in \mathbb{R}^3$ and $\bm{n}^*\in \mathbb{R}^3$: the position and normal vector along the desired curve in the robot base frame.
\item $e_x, e_z \in \mathbb{R}^3$: TCP $x$ and $z$-axis unit vector in the robot base frame. 
\item $J(\bm{q}) \bm{\dot{q}}=\bm{\nu}=\begin{bmatrix} \omega & v \end{bmatrix}^T$: Jacobian mapping between joint velocity $\bm{\dot{q}}$ and TCP spatial velocity $\bm{\nu}$.
\item $\lambda\in[0,\lambda_f]$: path length variable of the target curve from the start to the total path length by $\lambda_f$.
\item $\q0_i$: initial joint angle for robot $i$.
\item $\bm{v}^{\times}$: Skew symmetric matrix representing the cross product operation $\bm{v}\times$.  Let $\bm{v}=\begin{bmatrix} v_1 & v_2 & v_3 \end{bmatrix}^T$, then 
$$\bm{v}^{\times} := \begin{bmatrix} 0 & -v_3 & v_2 \\ v_3 & 0 & -v_1 \\ -v_2 & v_1 & 0 \end{bmatrix}.$$
\end{itemize}

\ifresubmissionred}\fi

\ifresubmissionred{\color{red}\fi

Given two robots $i=1,2$, with robot 1 designated as the spraying and robot 2 as the arm holding the target part, the robots need to traverse the 5-dof target curve parameterized by its position $\bm{p}^*(\lambda)$ and outward surface normal $\bm{n}^*(\lambda)$.

\ifresubmissionred}\fi

The dual arm trajectory optimization problem may be stated as maximizing the {\em constant} relative path speed subject to the dual arm kinematics and robot joint velocity and acceleration constraints:
\\ {\bf P1: Exact Dual-Arm Path Tracking Optimization }
\\
{\em
Given $\dot \lambda = \mu$, $\mu$ is a constant, $\lambda(0)=0$, $\lambda(t_f)=\lambda_f$, $t_f=\lambda_f/\mu$.  Find the configuration parameters $(\bm{q}\armsup1_0,\bm{q}\armsup2_0, \bm{T}^{(2)}_b)$ and robot joint velocities $(\bm{\dot q}\armsup1(\lambda),\bm{\dot q}\armsup2(\lambda))$, to maximize $\mu$ subject to
\begin{subequations}
    \begin{align}
        &\bm{p}\armsup 1 (\lambda)- \bm{p}\armsup 2 (\lambda) = \bm{R}\armsup2(\lambda) \bm{p}^*(\lambda) \label{eq:pathpos}\\
     &e_z\armsup 1(\lambda)= - \bm{R}\armsup 2(\lambda)\bm{n}^*(\lambda) \label{eq:pathnormal}\\
&\bm{\dot{q}}_{\min}\armsup i \preccurlyeq \bm{\dot q}\armsup i \preccurlyeq \bm{\dot{q}}_{\max}\armsup i, \,\,\,\,
\bm{\ddot{q}}_{\min}\armsup i \preccurlyeq \bm{\ddot q}\armsup i \preccurlyeq \bm{\ddot{q}}_{\max}\armsup i
    , \,\,\, {\displaystyle i=1,2.}
    \label{eq:qddotlimit}
\end{align}
\end{subequations}
}
\smallskip
Denote the arm forward kinematics as  
$\bm{p}\armsup i(\lambda)=f_p\armsup i(\bm{q}\armsup i(\lambda),\bm{T}_b\armsup i)$, $\bm{R}\armsup i(\lambda)=f_R\armsup i(\bm{q}\armsup i(\lambda),\bm{T}_b\armsup i)$. 
Path tracking constraints are captured in Eq.~\eqref{eq:pathpos}--\eqref{eq:pathnormal} for 
\ifresubmissionred{\color{red}\fi
position (3-dof) and surface normal (2-dof). The two robots have a combined 12-dof in their joint variables. 
\ifresubmissionred}\fi

Eq.~
\eqref{eq:qddotlimit} are the robot joint velocity and acceleration limits, where $\preccurlyeq$ denotes elementwise inequality.  We also assume there is an initial ramp up path, so $\bm{\dot q}\armsup i(0)$ may be nonzero.
Since we require the perfect path following:
\begin{equation}
\begin{aligned}
   \dot \lambda &= \frac{d \norm{\bm{p}^*(\lambda)}}{dt}= \frac{(\p1-\p2)\tr (\bm{\dot p}\armsup 1 - \bm{\dot p}\armsup 2)}{\norm{\p1-\p2}}\\
   &= \frac{(\p1-\p2)\tr }{\norm{\p1-\p2}}
   (J_v\armsup 1 (\bm{q}\armsup1) \bm{\dot q}\armsup 1 - J_v\armsup 2(\bm{q}\armsup2) \bm{\dot q}\armsup 2).
 \end{aligned}
 \label{eq:lambdadot}
\end{equation}
Then for each choice of the configuration parameters and robot joint velocities, there is a corresponding feasible region in the $(\lambda,\dot\lambda)$ space corresponding to the constraints.
\ifshortversion
The minimum of the infeasible region is the maximum allowed constant path speed.
The goal of the optimization problem is then to find the parameters and robot joint trajectory so that the minimum of the infeasible region is as large as possible.  
\fi
In practice, some tolerance of the constraints is acceptable.  The optimization problem may be relaxed to the following form:
\\ {\bf P2: Relaxed Dual-Arm Path Tracking Optimization }
\\
{\em
Given $\dot\lambda=\mu(\lambda)$, $\lambda(0)=0$, $\lambda(t_f)=\lambda_f$, $\integral 0{t_f} \dot \lambda(\tau)\,d\tau = \lambda_f$.
\\ Find the configuration parameters $(\bm{q}\armsup1_0,\bm{q}\armsup2_0, \bm{T}\armsup2_b)$ and robot joint velocities $(\bm{\dot q}\armsup1(\lambda),\bm{\dot q}\armsup2(\lambda))$, $\lambda\in[0,\lambda_f]$, to maximize $\mu_{avg}$, the mean of $\mu(\lambda)$ subject to
\begin{subequations}
    \begin{align}
    &\sigma(\mu(\lambda))\le \epsilon_{\mu} \mu_{avg} \label{eq:unifspeed}\\
    &\norm{\bm{p}\armsup 1 (\lambda)- \bm{p}\armsup 2 (\lambda) - \bm{R}\armsup 2(\lambda) \bm{p}^*(\lambda)}\le \epsilon_p \label{eq:pathposbound}\\
     & \abs{\angle(e_z{{\armsup 1}}(\lambda), -\bm{R}\armsup 2(\lambda)\bm{n}^*(\lambda))}\le \epsilon_n 
     \label{eq:pathnormalbound}\\
     &\mbox{and joint velocity and acceleration bounds in 
     \eqref{eq:qddotlimit}} \nonumber
\end{align}
\end{subequations}
}
The path speed standard deviation is allowed to vary by $\epsilon_\mu$ (usually specified as a percentage) about the average path speed.  The position and normal tracking errors are allowed to vary by $\epsilon_p$ and $\epsilon_n$, respectively.  In this study, we choose $\epsilon_\mu=5\%$, $\epsilon_p =0.5$~\SI{}{mm}, $\epsilon_n=3^\circ$, based on typical industry specifications for cold spray coating of turbine blades. 

\begin{figure}[!ht]
    \centering
    \includegraphics[width=0.42\textwidth]{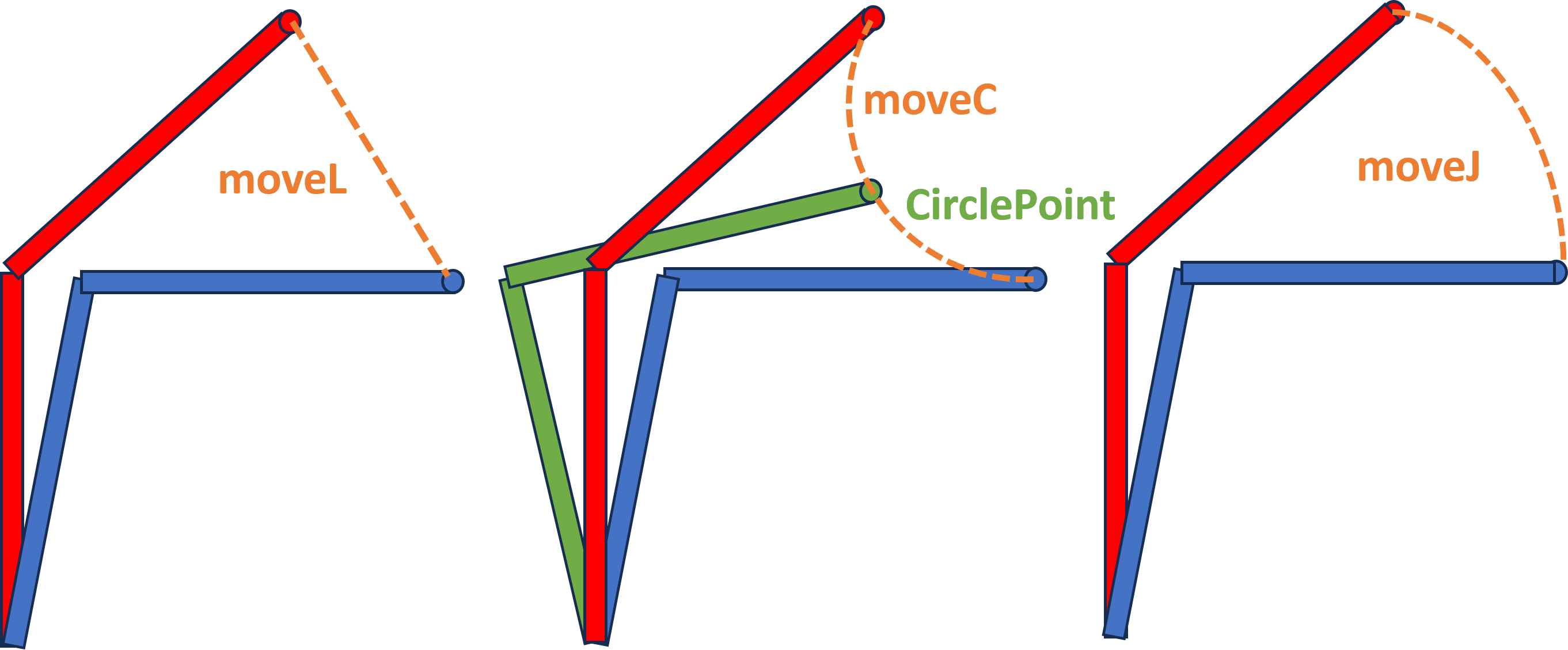}
    \caption{\small{Three types of robot motion primitives. {\tt moveL}: linear in Cartesian Space; {\tt moveC}: circular arc in Cartesian space, defined by start, end and a circlepoint; {\tt moveJ}: linear in joint space.}}
    \label{fig:motion_primitives}
\end{figure}

For industrial robots, there is an additional restriction. The path speed $\mu(\lambda)$ cannot be arbitrarily specified at each point on the path.  Instead, robot motion is specified in terms of waypoints connected by motion segments. As shown in Fig.~\ref{fig:motion_primitives}, every industrial robot provides three motion primitives for the motion segments:
\begin{itemize}
    \item {\tt moveL}: The robot motion is given by the straight line motion between the adjacent waypoints specified in the tool frames: $(\bm{p}_k,\bm{\bm{\beta}}_k)$ and $(\bm{p}_{k+1},\bm{\beta}_{k+1})$. 
    \item {\tt moveC}: The position portion of the robot motion is given by a circular arc in the tool frame.  The orientation portion is given by straight line interpolation in a specified parameterization.
    \item {\tt moveJ}: The robot motion is given by the straight line motion in the joint space between the adjacent waypoints specified in the joint space: $\bm{q}_k$ and $\bm{q}_{k+1}$.
\end{itemize}

Even though the motion primitives are universal, there are subtle differences from vendor to vendor, including target types (joint position or Cartesian pose with specified configuration), blending region specification (in radius or percentage) and speed (percentage or \SI{}{rad/s} or \SI{}{mm/s}). These vendor-specific arguments are taken care of in our robot drivers\footnote{https://github.com/rpiRobotics/abb\_motion\_program\_exec},\footnote{https://github.com/rpiRobotics/fanuc\_motion\_program\_exec} to parse into vendor-specific programming scripts, such that the optimization algorithm can interface directly.

A motion program for an industrial controller consists of a set of waypoints, indexed as $k=1,\ldots, K$ (with an initial condition as $k=0$), $K$ motion primitives connecting the waypoints, and the Cartesian path speed in the motion segment. 
\ifshortversion
To avoid high acceleration at the waypoints, one can specify a blending region. Different vendors may handle blending differently to smooth out the trajectory, such as blending in joint space directly or in Cartesian space.
\fi

Robot vendors use proprietary blending strategies and low-level motion control algorithms, so the exact behavior of the robot motion is not known {\em a priori}. In general, a large blending region tends to have smoother motion (path velocity is more uniform) but a larger error at the waypoint (distance between the motion and the waypoint). 

For dual-arm motion, one can specify the waypoints for both robots and they are synchronized based on the slower of the two path speeds so both robots can execute the motion segment in the same time period. The blending zone for dual-arm in synchronized motion is identical to the single-arm case, embedded in the motion primitive commands.

\section{Solution Approach}
\label{sec:approach}
\subsection{Summary Of Approach}
\label{sec:summary}
We decompose the problem 
\ifresubmissionred{\color{red}\fi
solution
\ifresubmissionred}\fi
into multiple steps.  The first step is to 
\ifresubmissionred{\color{red}\fi
find an approximate solution of the exact tracking 
\ifresubmissionred}\fi
optimization problem.  The second step is to approximate the optimal solution using the robot motion program consisting of motion primitives. The final step is to adjust the waypoints in the motion program to ensure the constraints are satisfied.  These steps are described further below:
\begin{itemize}
    \item {\em System Configuration Optimization}: For a given system configuration $(\q 1_0,\q 2_0, \bm{T}_b\armsup 2)$, we use Jacobian-based inverse kinematics for redundancy resolution to find the entire joint space trajectory under constraints \eqref{eq:pathpos}--\eqref{eq:qddotlimit}. A speed estimation model is used to find the maximum constant $\mu$, we then use differential evolution to find the best system configuration to maximize $\mu$.
    \item {\em Waypoint Optimization}: The solution in step 1 is not directly realizable using industrial robot motion programs.  Instead, we will approximate the target curve in terms of motion primitives to within a specified tolerance while using the smallest number of waypoints.  The reason is that waypoints tend to compromise performance due to the need to blend motion segments.  We apply greedy search in this step by choosing the best motion primitive type for each motion segment.  We then use the simulator from the robot vendor to check the speed uniformity constraint \eqref{eq:unifspeed}.  The robot velocity and acceleration constraints are typically already enforced in the simulator.  The path speed is then lowered from step 1 until the speed uniformity constraint is satisfied.
    \item {\em Waypoint Iteration}: The final step is to adjust the waypoints directly to improve the path constraints \eqref{eq:unifspeed}--\eqref{eq:pathnormalbound}.
    This step is performed first in simulation and continued with the physical robots if available.    
\end{itemize}

\subsection{System Configuration Optimization}
\label{sec:config}

\ifresubmissionred{\color{red}\fi

Finding the joint path of the robots, $(\bm{q}\armsup1(\lambda),\bm{q}\armsup2(\lambda))$ subject to the path tracking constraint \eqref{eq:pathpos}-\eqref{eq:pathnormal} is a redundancy resolution problem.  We apply the standard Jacobian-based inverse kinematics \cite{chiaverini1997singularity} along the specified path, with the previous path point as the initial guess. The $6\times12$ relative Jacobian $\bar J(\bm{q})\!\!=\!\!\ma{cc}\bar J\armsup1(\q1) & \bar J\armsup2(\q2)\ema$ is obtained by differentiating \eqref{eq:pathpos}-\eqref{eq:pathnormal}, where 
\begin{subequations}
\begin{align}
\!\!\!\!\bar J\armsup1(\q1) & = \ma{c}\J1_v(\q1) \\ {e_z\armsup1}^\times \J1_\omega(\q1) \ema \\
\!\!\!\!\bar J\armsup2(\q2) & = \ma{c}-\J2_v(\q2) +(\bm{R}\armsup 2 \bm{p}^*)^\times \J2_\omega(\q2) \\ (\bm{R}\armsup 2 \bm{n}^*)^\times \J2_\omega(\q2)\ema\!\!.
\end{align}
\end{subequations}

\begin{figure}[!ht]
    \centering
    \includegraphics[width=0.35\textwidth]{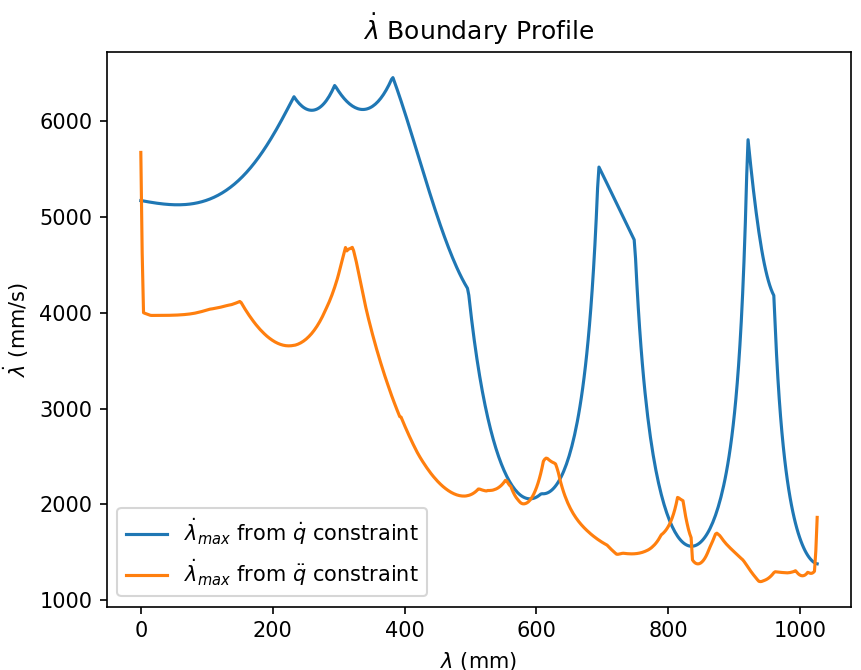}
    \caption{\small{\ifresubmissionred\color{red}\fi Curve 1 relative trajectory traversal speed boundary profile based on both arms' joint velocity (blue) and configuration-dependent acceleration constraints (orange). The minimum of both establishes the feasible $\ldot_{\max}$ of the relative trajectory.\ifresubmissionred\color{black}\fi}}
    \label{fig:lamdot_profile}
\end{figure}

Given the joint path of the robots, $(\bm{q}\armsup1(\lambda),\bm{q}\armsup2(\lambda))$, corresponding to the specified relative path, 
the maximum constant path speed subject to the constraints \eqref{eq:qddotlimit} may be found exactly. 
Along the joint path, we have 
\begin{subequations}  
\begin{align}
    \bm{\dot q}\armsup i & = {\bm{q}\armsup i}'(\lambda)\ldot  \\
    \bm{\ddot q}\armsup i & = {\bm{q}\armsup i}''(\lambda)\ldot\sq + {\bm{q}\armsup i}'(\lambda)\lddot.
\end{align}
\end{subequations}
Substituting into \eqref{eq:qddotlimit}, we have 
\begin{subequations}
\begin{align}
& \bm{\dot{q}}_{\min}\armsup i \preccurlyeq {\bm{q}\armsup i}'(\lambda)\ldot \preccurlyeq \bm{\dot{q}}_{\max}\armsup i
\label{eq:ldot_first}
\\
& \bm{\ddot{q}}_{\min}\armsup i \preccurlyeq {\bm{q}\armsup i}''(\lambda)\ldot\sq+{\bm{q}\armsup i}'(\lambda)\lddot \preccurlyeq \bm{\ddot{q}}_{\max}\armsup i.
\label{eq:ldot_second}
\end{align}
\label{eq:pathvelaccel}
\end{subequations}
\!\!From the uniform speed requirement, we have $\ddot\lambda=0$. Then we can easily find the largest $\dot\lambda$ that satisfies each of the constraints: \eqref{eq:ldot_first} from $\bm{\dot{q}}\armsup i$  and \eqref{eq:ldot_second} from $\bm{\ddot{q}}\armsup i$. Fig.~\ref{fig:lamdot_profile} shows the $\ldot$ maximum boundary for each of these constraints.  The lower of the two is the constraint $\dot\lambda_{\max}$.  The minimum of the path speed boundary over the entire path is the maximum uniform path speed $\mu=\min_\lambda \dot\lambda_{\max}(\lambda)$.

Industrial robot vendors usually provide joint velocity limits in the data sheets, but the acceleration limit is undisclosed due to the torque limit and dependence on the robot dynamics, load, and robot configuration. We have used an experimental procedure to determine the robot acceleration limit in various poses
\ifresubmissionred{\color{red}\fi
as described in \cite{icra_singlearm}.
\ifresubmissionred}\fi
The acceleration identification routine is available in open-source on our GitHub repository\footnote{https://github.com/rpiRobotics/Robot\_Acceleration\_Identification}.

\begin{figure}[!ht]
     \centering
         \includegraphics[width=.185\textwidth]{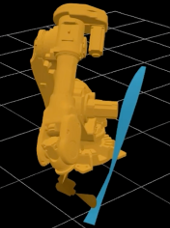}
         \includegraphics[width=.27\textwidth]{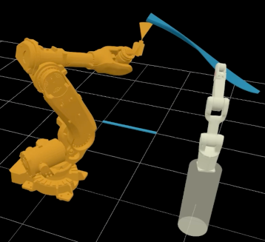}         
     \caption{\small{Optimized configuration for the single-arm case \cite{icra_singlearm} vs. the dual-arm case in this paper.}}
     \label{fig:curve2_optimized_poses}
\end{figure}

\ifresubmissionred}\fi
With the estimated traversal speed $\mu$ for 
\ifresubmissionred{\color{red}\fi
a given set of configuration parameters, 
\ifresubmissionred}\fi
we next apply the differential evolution algorithm \cite{diff_evo} 
to find the global optimized configuration to maximize the path traversal speed. Differential evolution is a global optimization evolution algorithm that involves mutation (generating new solutions by combining existing ones), crossover (mixing mutant solutions with current solutions), and selection (choosing the better solution for the next generation). The initial guess is the single-arm baseline, as described in \cite{icra_singlearm}, where Robot 2 just holds the part stationary.  
In our testbed, Robot 1 is positioned flush on the floor and Robot 2 is mounted on a mobile pedestal.  Hence, the relative pose of $\bm{T}_b\armsup 2$ is 
\ifresubmissionred{\color{red}\fi
a planar transformation with 3 parameters: two distances and a relative rotation angle.
\ifresubmissionred}\fi
The target trajectory is mounted close to the part center of mass so as to minimize the torque generated on the robot wrist.  Together with $(\q1_0,\q2_0)$, there are 15 parameters to optimize in this step.   
The optimized dual-arm and single-arm system configurations for a sample turbine blade leading edge are shown in Fig.~\ref{fig:curve2_optimized_poses}.

\begin{figure}[!ht]
    \centering
    \includegraphics[width=0.33\textwidth]{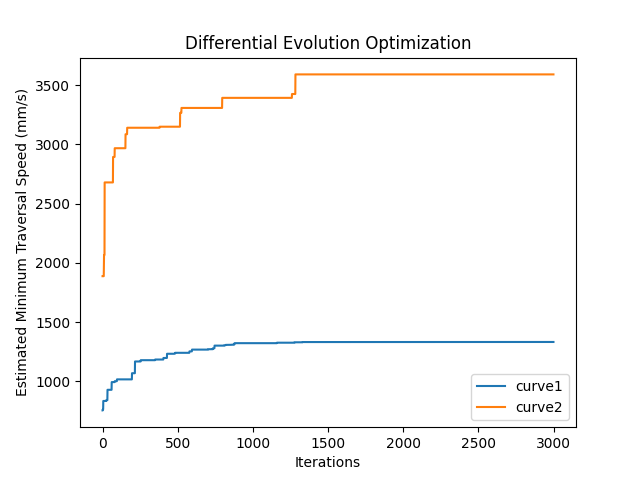}
    \caption{\small{Differential evolution optimization progression for the estimated path traversal speed $\mu$ for the two test curves in Section~\ref{sec:curves}. The final outputs are the complete robot joint trajectories and the robot base relative pose.}}
    \label{fig:diffevo_result}
\end{figure}

\ifresubmissionred{\color{red}\fi
The progression of the differential evolution is shown in Fig.~\ref{fig:diffevo_result}.
\ifresubmissionred}\fi
The maximum number of iterations is set to 3000.  The optimal path speeds for the two test curves are 1332~\SI{}{mm/s} and 3589~\SI{}{mm/s},
\ifresubmissionred{\color{red}\fi
improving from the starting points 680~\SI{}{mm/s} and 1757~\SI{}{mm/s}, respectively.

\begin{figure}[ht]
     \centering
     \begin{subfigure}[b]{0.239\textwidth}
         \centering
         \includegraphics[width=\textwidth]{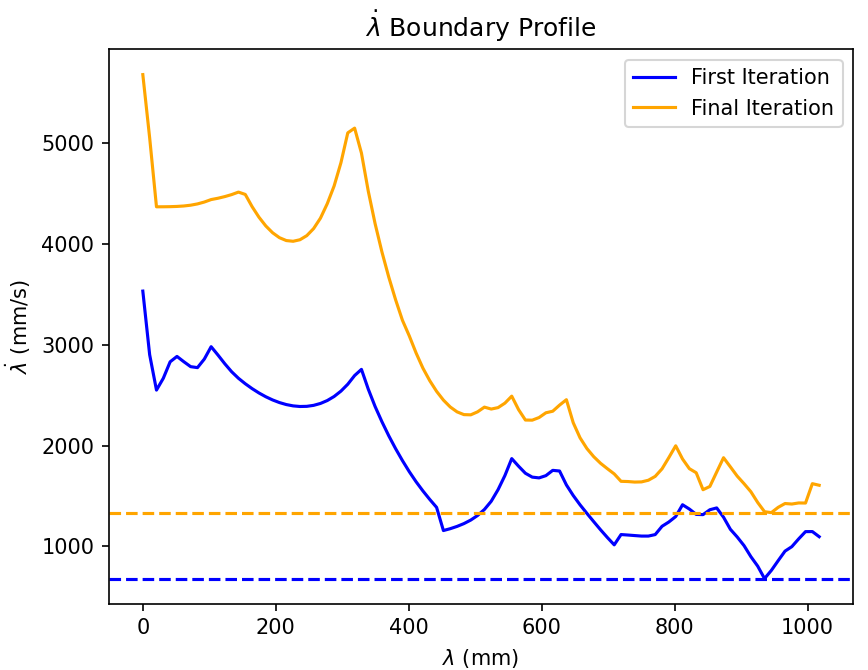}
         \caption{Curve 1}
         \label{fig:lamdot_optimization_curve1}
     \end{subfigure}
     \begin{subfigure}[b]{0.239\textwidth}
         \centering
         \includegraphics[width=\textwidth]{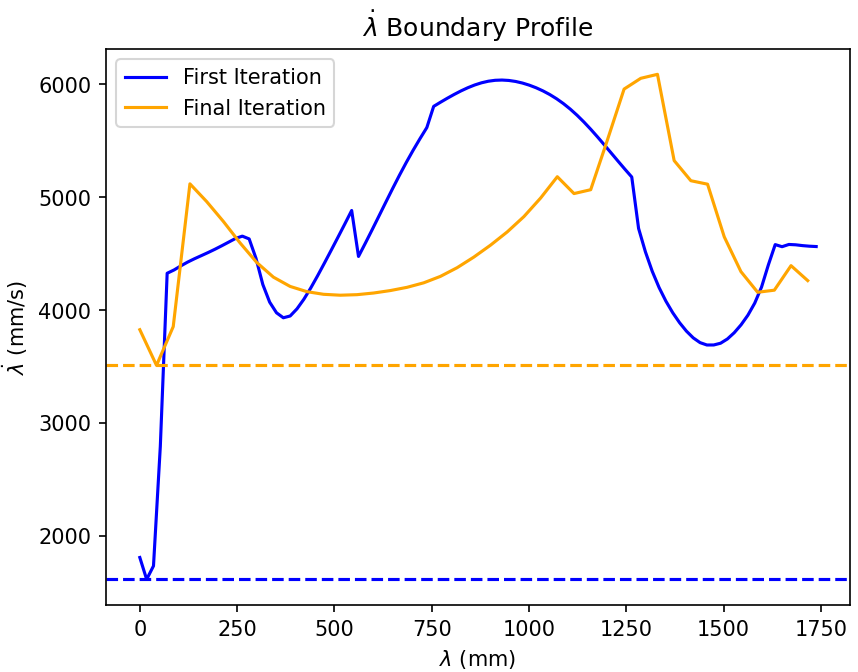}
         \caption{Curve 2}
         \label{fig:lamdot_optimization_curve2}
     \end{subfigure}
        \caption{\small{$\ldot$ optimization boundary profile comparison of first and final iterations in differential evolution optimization for Curve 1 and Curve 2, where the $\min_\lambda \dot\lambda_{\max}(\lambda)$ is chosen as the commanded relative traversal speed $\mu$.}}
        \label{fig:lamdot_optimization_comparison}
     \vspace{-1em}
\end{figure}

The $\ldot_{max}$ boundary profile comparison between the initial step and final iteration in the differential evolution is shown in Fig.~\ref{fig:lamdot_optimization_comparison} for both test curves.
\ifresubmissionred}\fi
Due to the unknown vendor-specific controller behaviors such as blending strategy and inner loop control settings, such high speeds are likely not achievable. We will discuss the implementation on the physical testbed in Section~\ref{sec:results}.

\vspace{-1em}

\leavevmode

\leavevmode

\def\qq{\bm{q}}
\def\qqdot{\bm{\dot q}}
\def\qqddot{\bm{\ddot q}}
\def\pp{\bm{p}}
\subsection{Motion Primitive Planning with Greedy Fitting}
\label{sec:fitting}

\begin{figure}[ht]
     \centering
     \begin{subfigure}[b]{0.22\textwidth}
         \centering
         \includegraphics[width=\textwidth]{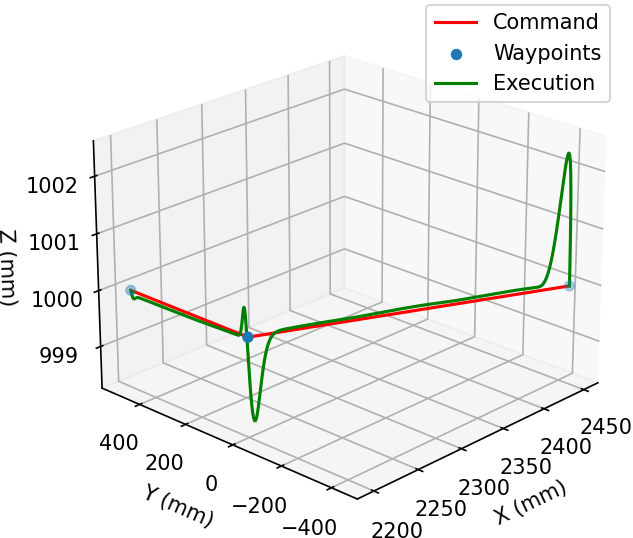}
         \caption{Two {\tt moveL} blending}
         \label{fig:blending_log}
     \end{subfigure}
     \begin{subfigure}[b]{0.259\textwidth}
         \centering
         \includegraphics[width=\textwidth]{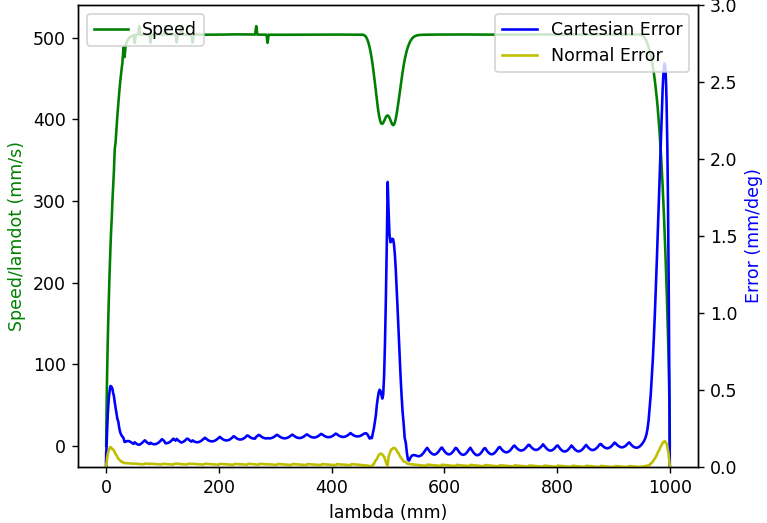}
         \caption{Segment Blending Performance}
         \label{fig:blending_traj}
     \end{subfigure}
        \caption{\small{ABB6640 30$^\circ$ slope {\tt moveL} blending at 500~\SI{}{mm/s} and blending zone of 1~\SI{}{mm}. The blending in joint trajectory causes out-of-plane motion with large error and slow down.}}
        \label{fig:blending_exp}
\end{figure}

The solution from Section~\ref{sec:config} will provide a feasible trajectory.  However, it cannot be directly implemented using standard industry robot controllers as discussed in Section~\ref{sec:summary}. Current industry practice involves using {\tt moveL} to interpolate the robot trajectory. 
\ifresubmissionred{\color{red}\fi
Gleeson et al. have introduced a pipeline to parse joint space trajectories to RAPID programs for ABB robots with linear segments \cite{robot_code_iter}.
\ifresubmissionred}\fi
While it's possible to densely interpolate the trajectory directly with {\tt moveL} or {\tt moveJ} to achieve the closest motion primitive command, robot controllers will prevent close waypoints programming due to various reasons including overlapping blending zone, infeasible corner path, or a hard stop at certain waypoints with failed/interference blending. Furthermore, segments blending also compromise performance as shown in Fig.~\ref{fig:blending_exp}, and different vendors have different blending strategies. Therefore, we want to use the smallest number of waypoints with all three motion primitives that can fit a given curve.

In \cite{icra_singlearm}, we introduced the greedy fitting algorithm for the single-arm trajectory planning case. Motion primitive trajectory greedy fitting starts at the initial point and looks for the longest possible segment under the specified accuracy threshold with {\tt moveL/moveC/moveJ} with unconstrained regression using bisection search. This step continues on the subsequent trajectory with continuity constraints. 
 
For multi-robot control, there are various ways to command the robots. Similar to a single-arm control, it is possible to set up our drivers on two socket-connected computers talking to two individual IRC5 controllers respectively, or set up the drivers on the same computer as a centralized control computer. Both ways are feasible for slow-motion applications with more accuracy tolerance. However, for high-speed and high-accuracy requirements, we decided to take advantage of the vendor-proprietary controller, Multimove by ABB, MultiArm by FANUC and Coordinated Control Function by Motoman to achieve a low-level synchronized clock. Therefore, dual-arm motion command consists of two sets of motion primitives with shared waypoint indexing, meaning the motion of the two robots will be synchronized to reach the individual commanded waypoint at the same time.  

\begin{figure}[!ht]
    \vspace{0.8em}
    \centering
     \begin{subfigure}[b]{0.239\textwidth}
         \centering
         \includegraphics[width=\textwidth]{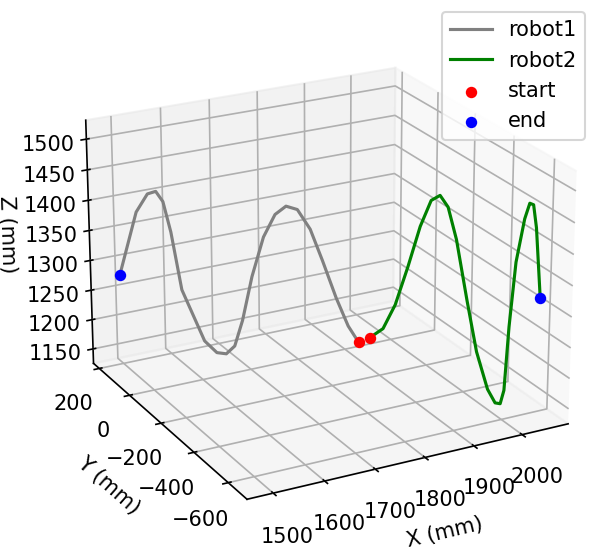}
         \caption{Dual arm Cartesian trajectories.}
         \label{fig:greedy_dual_separated}
     \end{subfigure}
     \begin{subfigure}[b]{0.24\textwidth}
         \centering
         \includegraphics[width=\textwidth]{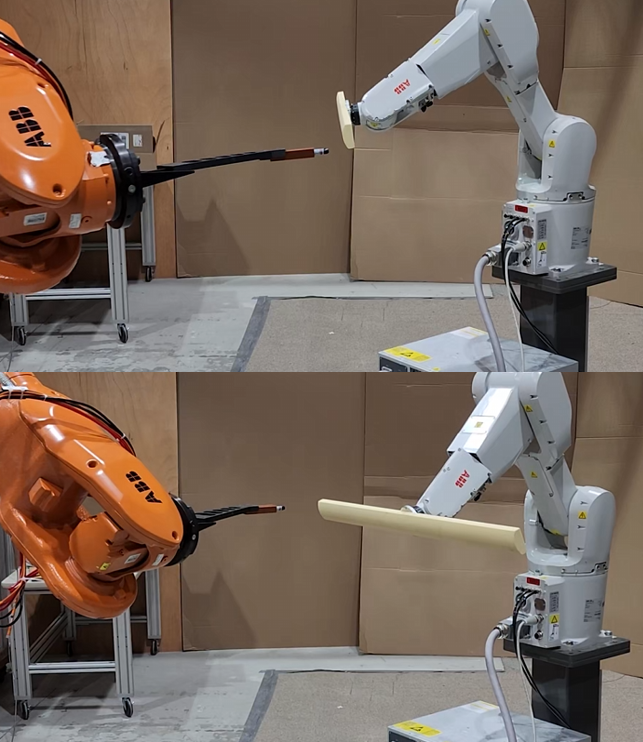}
         \caption{Curve 1 start and end poses.}
         \label{fig:greedy_dual_projection_partial}
     \end{subfigure}
    \caption{\small{
    Dual arm Greedy Fitting on Curve 1: each robot has its own fitted motion primitive sequences. Starting from the red initial point, each pair motion segment is executed synchronously and finally arrives at the blue end point.}}
    \label{fig:greedy_fitting}
\end{figure}

For dual-arm applications, the greedy fitting algorithm is modified as shown in Fig.~\ref{fig:greedy_fitting}. Each robot has its own trajectory, and {\tt moveL/moveC/moveJ} regression is performed on its Cartesian/joint space trajectory. Starting from the initial point, the greedy fitting performs unconstrained {\tt moveL/moveJ/moveC} regression for robot 1 and robot 2 trajectories respectively. But the threshold is checked based on relative trajectory error, meaning the fitted motion primitive segment for robot 1 and robot 2 will be converted in robot 2's TCP frame and calculates the tracking error to the original curve segment.
We use bisection search to identify the furthest shared waypoint index, such that the relative error is under the specified threshold. And this process continues on both robot arm's subsequent trajectories. 
\ifresubmissionred{\color{red}\fi
For Curve 1, the motion primitive sequences of Robot 1 and Robot 2 are $J^3L^1J^5L^1J^3L^3J^2L^2$ and  $J^1L^2J^2L^4J^2L^4J^2L^3$, respectively.  The letters $L$, $C$, $J$ correspond to the {\tt moveL}, {\tt moveC} and {\tt moveJ} primitives, and the superscript means the number of segments with the same primitive. For Curve 2, the motion primitive sequences of Robot 1 and Robot 2 are  $L^{12}J^2L^6J$ and $L^{12}J^2C^2J^5$, respectively.
\ifresubmissionred}\fi
The motion primitive sequence is extended at the first and last motion segments for the robots to accelerate/decelerate.  Note that the number of motion segments is the same for the two robots since the robot controllers synchronize at the waypoints.


\subsection{Waypoint Adjustment}
\label{sec:waypoint_adjust}

The waypoints and motion primitives from Section~\ref{sec:fitting} together with the optimized path speed $\mu$ from Section~\ref{sec:config} may be incorporated into the robot motion program and applied to the robot.  However, invariably, the performance specification \eqref{eq:unifspeed}
--\eqref{eq:pathnormalbound} will not be satisfied.  This is caused by the effect of speed variation due to blending and the inherent performance limitation of the robot (velocity and acceleration limits) and robot controller. This performance degradation is dependent on the robot payload and configuration and is not possible to predict accurately without additional information from the robot vendor.

We apply an iterative approach to adjust the waypoints in the motion program based on the measured tracking error, first in simulation and then from physical robots, to improve the tracking accuracy.  For each path speed $\dot\lambda=\mu$, we use waypoint iteration to reduce the tracking error.  The path speed $\mu$ is reduced until all constraints \eqref{eq:unifspeed}--\eqref{eq:pathnormalbound} are satisfied.

From the single-arm setup, we have developed a proportional adjustment strategy based on reinforcement learning derived policy \cite{aim2023}, where all waypoints are adjusted in the direction of the tracking error vector. This strategy reduces the tracking error but may not meet the accuracy requirement.  We follow up with a multi-peak gradient descent approach to adjust only waypoints with large tracking errors, using the gradient with respect to the waypoints and their immediate neighbors calculated based on an assumed blending using cubic splines. 
This process is summarized below:
\begin{itemize}
    \item Proportional Adjustment: Adjust all waypoints for both robots in the relative error direction. For the $k$th waypoint and $i$th robot, 
    \begin{equation}
        \begin{aligned}
            &\bm{p}^{(1)}_k \leftarrow \bm{p}^{(1)}_k - \gamma {\bm{e}_p}_k ,\,\,\,
            \bm{p}^{(2)}_k \leftarrow \bm{p}^{(2)}_k + \gamma {\bm{e}_p}_k \\
            &\bm{R}^{(1)}_k \!\! \leftarrow {\sf R}(e^\perp_k,\gamma \theta_k) \bm{R}^{(1)}_k  ,\,   
            \bm{R}^{(2)}_k \!\! \leftarrow {\sf R}(e^\perp_k,-\gamma \theta_k) \bm{R}^{(2)}_k
        \end{aligned}
    \end{equation}
    where $\bm{e}_p$ is the position error from \eqref{eq:pathposbound}, $\theta$ is the angular error from \eqref{eq:pathnormalbound}:
\begin{equation}
\begin{aligned}
\bm{e}_p = {\bm{p}\armsup 1 - \bm{p}\armsup 2  - \bm{R}\armsup 2 \bm{p}^*}, \, \theta = \!\!{\angle(e_z{{\armsup 1}}, -\bm{R}\armsup 2 \bm{n}^*)}
\end{aligned}
\end{equation}
$e^\perp$ is the normalized vector of ${e_z\armsup1}^\times\bm{R}^{(2)}\bm{p}^*$, ${\sf R}$ denotes the rotation matrix for a given axis and angle, $\gamma$ is the step size, $\bm{p}^*_k$ is the point on the specified curve closest to $\bm{p}_k$.
\item Multi-peak Gradient Descent: 
After the proportional adjustment no longer reduces the worst case tracking error,
we target error peaks in the region where the error exceeds the tolerance.  Let $p_{peak}$ be a peak error point on the trajectory. 
Define its error as $\bm{p}_{err}=\bm{p}_{peak}-\bm{p}_{peak}^*$, where $\bm{p}_{peak}^*$ is the closest point on the desired relative trajectory. Identify the closest 3 waypoints in the relative commanded trajectory, call them $\bm{p}_{wp_i}$, $i=1,\ldots,3$, corresponding to two sets of synchronized waypoints $\bm{p}^{(1)}_{wp_i},\bm{p}^{(2)}_{wp_i}$, for each arm. Define $\bm{\hat{p}}_{peak}$ as the closest point to ${p}_{peak}$ in a cubic spline interpolated predicted trajectory. Then calculate the gradient numerically $\frac{\delta {\bm{\hat{p}}_{peak}}}{\delta {\bm{p}_{wp}}_i}$, which is an estimate of the actual error gradient on the executed trajectory.
We can then estimate the error gradient $\frac{\delta \bm{e}_p}{\delta {\bm{p}_{wp}}_i}$ numerically, and use it to update the waypoints by gradient descent with adaptive step size. The same scheme is used for the orientation waypoint adjustment.

    
\end{itemize}

\begin{figure}[ht]
     \centering
     \begin{subfigure}[b]{0.4\textwidth}
         \centering
         \includegraphics[width=\textwidth]{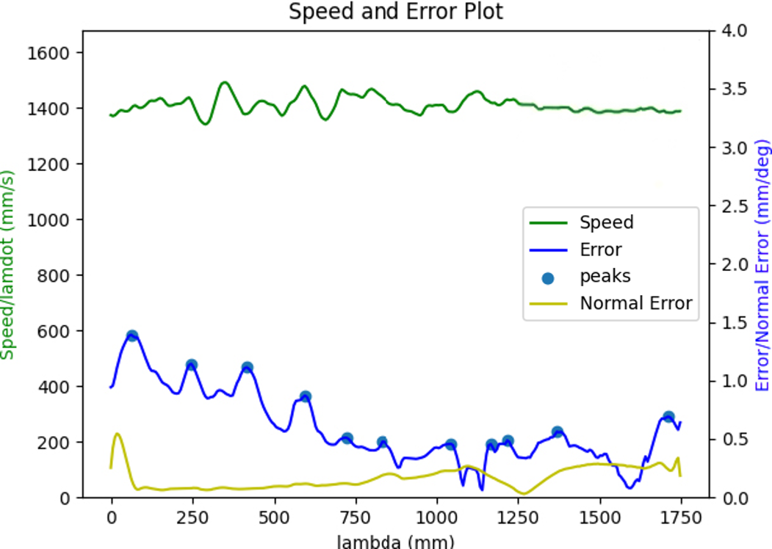}
         \caption{\small{Execution results with initial waypoints}}
         \label{fig:initial_execution}
     \end{subfigure}
     \begin{subfigure}[b]{0.4\textwidth}
         \centering
         \includegraphics[width=\textwidth]{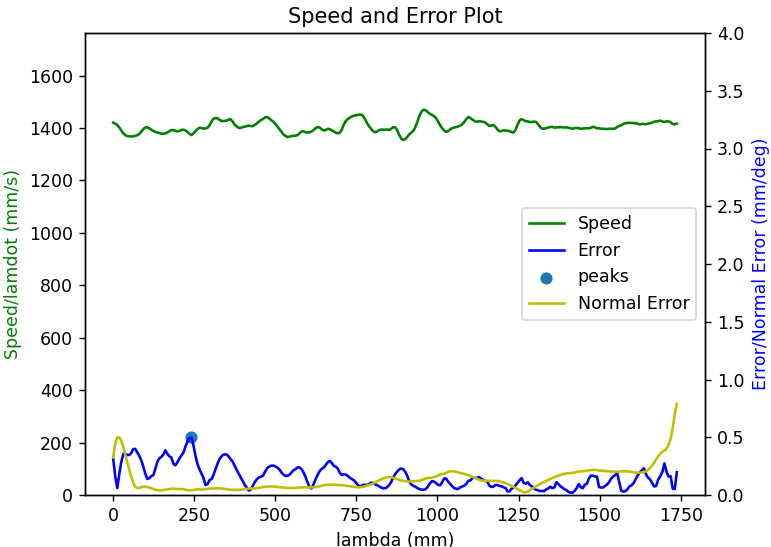}
         \caption{\small{Results with updated waypoints after five iterations}}
         \label{fig:updated_waypoints}
     \end{subfigure}
        \caption{\small{Execution results comparison (path speed, position error, normal error) initial waypoints vs.~waypoints after five iterations. Peak position errors beyond the 0.5~\SI{}{mm} threshold are highlighted.}}
        \label{fig:waypoint_progression}
\end{figure}

Fig.~\ref{fig:waypoint_progression} shows the waypoint update progress for the test curve 2 on the physical robot trajectory execution, with the initial execution command from the best simulation using the robot vendor simulator.  After five iterations, the relative tracking error with the updated waypoint command is below the 0.5~\SI{}{mm} requirement.






\section{Implementation and Evaluation}
\subsection{Test curves and Baseline}
\label{sec:curves}

\begin{figure}[!ht]


      \begin{subfigure}[b]{0.239\textwidth}
         \centering
         \includegraphics[width=\textwidth]{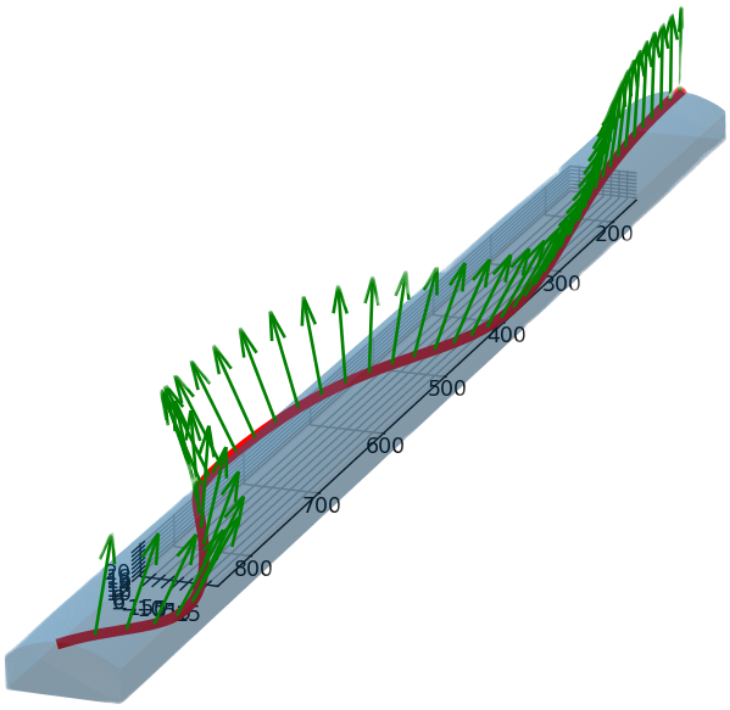}
         \caption{\small{Curve 1: frequency-changing sinusoidal curve on a parabolic surface.}}
         \label{fig:curve1}
     \end{subfigure}
     \begin{subfigure}[b]{0.239\textwidth}
         \centering
         \includegraphics[width=\textwidth]{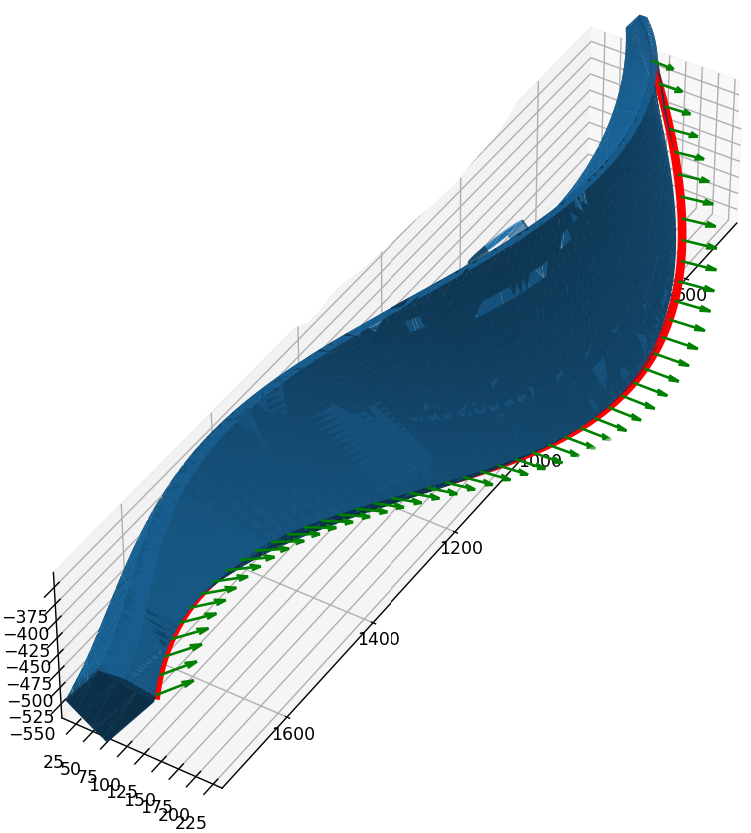}
         \caption{\small{Curve 2: generic Fan Blade leading edge curve.}}
         \label{fig:curve2}
     \end{subfigure}
    \caption{\small{Representative two 5-dof spatial curves.}}
    \label{fig:curves}
\end{figure}

Based on industry needs, we choose two representative test curves as shown in Fig.~\ref{fig:curve1} to evaluate our algorithms. Curve~1 is a multi-frequency spatial sinusoidal curve on a parabolic surface. This curve is representative of a high curvature spatial curve. Curve~2, shown in Fig. \ref{fig:curve2}, is extracted from the leading edge of a generic fan blade model, which is a typical case for cold spraying applications in industry.

Based on the current industry practice, 
we developed a baseline approach for a single arm tracking a stationary cure as  a benchmark 
\cite{icra_singlearm}. The target curve is positioned at the center of the robot workspace to avoid singularities. The redundant tool-$z$ orientation is addressed by aligning the tool $x$-axis with the motion trajectory. The selection of the robot arm pose is chosen by maximizing the manipulability. This baseline utilizes equally spaced {\tt moveL} segments to interpolate the trajectory. The performance is determined by identifying the maximum allowable path speed that meets the criteria for traversal speed consistency and tracking accuracy specified in Section \ref{sec:problem_formulation}.  This approach will also serve as the baseline for the dual-arm case, with the second arm remaining stationary.

\subsection{Simulation and Physical Setup}

\ifshortversion

Dual-arm coordination with high precision requires a synchronized clock for both arms. We used MultiMove Slave Drive Module 
with IRC5 controller 
with RobotWare 6.13 to drive both ABB arms in our testbed synchronously. 

\fi
We have developed a Python driver to interface to the IRC5 controller and RobotStudio virtual controller to execute motion commands directly. For all motion commands, we start with a blending zone of 10~\SI{}{mm} and gradually increase it until the speed profile converges such that the robot does not slow down around waypoints due to blending.  

Fig.~\ref{fig:curve2_abb} shows the experiment setup for Curve 2, with the ABB6640 robot holding a mock spray gun and the ABB1200 robot holding the part to be machined. The joint trajectory is recorded through the robot controller at 250~\SI{}{Hz} for all 12 joints. Robot forward kinematics is used to compute the TCP locations and the relative trajectory.





According to the manuals of ABB6640 and ABB1200, the path repeatability is up to 1.06~\SI{}{mm} and 0.07~\SI{}{mm} respectively. Therefore, we run the robot 5 times for each set of the motion primitive commands and take the interpolated average of 5 recorded trajectories as the execution trajectory. 

\subsection{Results}
\label{sec:results}

We have applied our methodology using both RobotStudio simulation and physical robots, and compared with the baseline and optimized single-arm approach as in \cite{icra_singlearm}.  The results are summarized below.


\begin{table}[!ht]
\centering
\begin{tabular}{|c|p{1.2cm}|p{1.cm}|p{0.8cm}|p{1.3cm}|} 
 \hline
 Curve 1 & $max\,\, \norm{\bm{e}_p}$ & $max\, \theta$ & $\mu_{avg}$ & $\sigma(\mu)/\mu_{avg}$ \\ 
[0.2ex]  & \SI{}{\mm} & $^{\circ}$ & \SI{}{mm/s} & \% \\
 \hline\hline
 Baseline (sim)&            0.49 & 0.18 & 124.01 & 0.82     \\ 
 Baseline (real)&           0.46 & 0.21 & 103.11 & 0.82     \\
 Single Opt (sim)&          0.48 & 0.57 & 399.81 & 1.62     \\   
 Single Opt (real)&         0.41 & 0.33 & 395.72 & 2.41     \\
 Dual Opt (sim)&            0.42 & 1.49 & 550.73 & 4.13   \\ 
 Dual Opt (real)&           0.41 & 2.29 & 451.52 & 3.74   \\[0.5ex] 
 \hline
\end{tabular}
\begin{tabular}{|c|p{1.2cm}|p{1.cm}|p{0.8cm}|p{1.3cm}|} 
 \hline
 Curve 2 & $max\,\, \norm{\bm{e}_p}$ & $max\, \theta$ & $\mu_{avg}$ & $\sigma(\mu)/\mu_{avg}$ \\
 \hline\hline
 Baseline (sim)&            0.42 & 0.26 & 406.10    & 0.23     \\ 
 Baseline (real)&           0.47 & 0.28 & 299.97    & 0.79     \\
 Single Opt (sim)&          0.42 & 1.09 & 1207.45   & 0.95   \\
 Single Opt (real)&         0.46 & 1.15 & 1197.43   & 1.11    \\
 Dual Opt (sim)&            0.47 & 0.69 & 1705.26   & 1.10   \\ 
 Dual Opt (real)&           0.50 & 0.80 & 1404.87   & 1.52   \\[0.5ex] 
 \hline
\end{tabular}
\caption{\small Results for RobotStudio Simulation and Physical Robot Experiments. $\mu_{avg}$ denotes average path speed, $\sigma(\mu)$ denotes standard deviation of speed along path.  \textbf{Baseline}: Current industry practice. \textbf{Single Opt}: Results after optimization with single-arm (ABB6640) \cite{icra_singlearm} \textbf{Dual Opt}: Results for using the approach in this paper for the dual-arm system of ABB6640 and ABB1200.}
\label{table:results}
\end{table}
The final optimized dual-arm results show 344\% (sim) 337\% (real) speed increase for Curve 1 and 319\% (sim) and  368\% (real) speed increase for Curve 2 over baseline while keeping the tracking accuracy within the requirement. The dual-arm results also show 38\% (sim) 14\% (real) speed increase for Curve 1 and 41\% (sim) and  17\% (real) speed increase for Curve 2 over single-arm optimized results. Since we used a much smaller robot as the second arm, the dual-arm results are not significantly improved over the optimized single-arm case. Since both robots are running on the shared master controller (IRC5), faster commanded speed will throw Corner Path Failure error from motion segment blending, which prevents the robots from achieving their theoretical performance. We are also seeing the performance drop in reality over simulation. However, this systematic approach is applicable and optimized to all 5-dof trajectory traversal applications with dual-arm configuration. 

\ifresubmissionred{\color{red}\fi
\subsection{Implementation for Dual-arm FANUC Robots}

\begin{figure}[!ht]
    \vspace{0.8em}
    \centering
\includegraphics[width=0.4\textwidth]{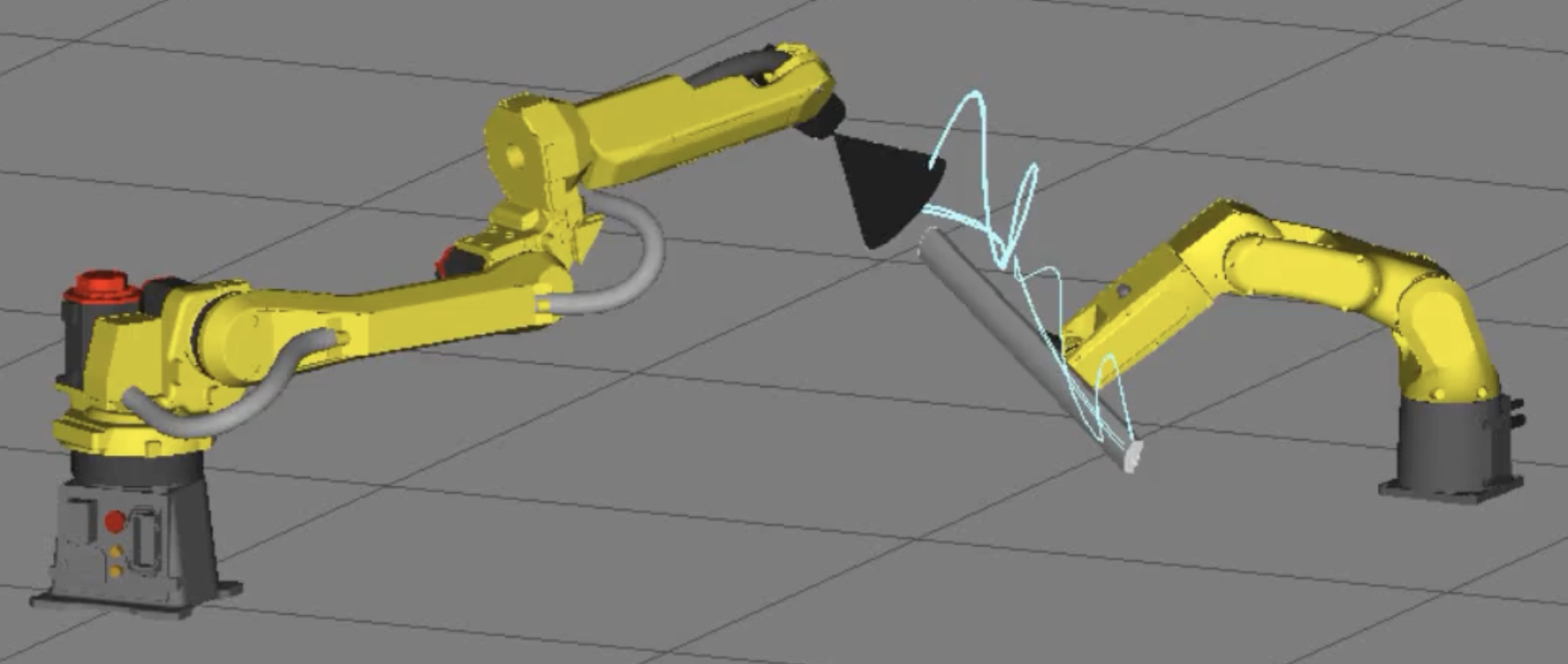}
    \caption{\small FANUC M10iA (left) + LR Mate 200iD (right) Dual-arm Simulation in RoboGuide for Curve 1.}
    \label{fig:fanuc_dual}
    \vspace{-0.3em}
\end{figure}

To demonstrate the applicability of the proposed methodology to other industrial robots, we evaluate the approach on FANUC M10iA and LR~Mate 200iD using the FANUC simulation program RoboGuide as shown in Fig.~\ref{fig:fanuc_dual}.  The result is summarized in Table~\ref{table:fanuc_sim_results}. The RoboGuide simulation has shown 680\% speed increase for Curve 1 and 587\% speed increase for Curve 2 over baseline; and compared with single-arm optimized results, it has 32\% speed increase for Curve 1 and 45\% speed increase for Curve 2.  We expect the performance to degrade on the physical robots, but still with substantial improvement over the baseline.

\begin{table}[!ht]
\centering
\begin{tabular}{|c|p{1.2cm}|p{1.cm}|p{0.8cm}|p{1.3cm}|} 
 \hline
 Curve 1 & $max\,\, \norm{\bm{e}_p}$ & $max\, \theta$ & $\mu_{avg}$ & $\sigma(\mu)/\mu_{avg}$ \\ 
[0.2ex]  & \SI{}{\mm} & $^{\circ}$ & \SI{}{mm/s} & \% \\
 \hline\hline
 Baseline         &            0.14 & 0.10 & 55.54 & 4.00     \\ 
 Single Opt&           0.48 & 0.63 & 327.15 & 4.81     \\   
 Dual Opt&             0.41 & 2.86 & 433.55 & 4.64  \\ 
 \hline
\end{tabular}
\begin{tabular}{|c|p{1.2cm}|p{1.cm}|p{0.8cm}|p{1.3cm}|} 
 \hline
 Curve 2 & $max\,\, \norm{\bm{e}_p}$ & $max\, \theta$ & $\mu_{avg}$ & $\sigma(\mu)/\mu_{avg}$ \\

 \hline\hline
 Baseline         &            0.25 & 0.34 & 144.15    & 4.95     \\ 
 Single Opt&           0.39 & 1.50 & 679.12  & 4.95     \\   
 Dual Opt&             0.44 & 2.26 & 980.08  & 1.33  \\ 
 \hline
\end{tabular}
\caption{\small Performance Comparison for FANUC Robots in RoboGuide Simulation}
\label{table:fanuc_sim_results}
 \vspace{-.2in}
\end{table}

\ifresubmissionred}\fi


The entire workflow of the motion optimization process is integrated into a Python Qt user interface, as documented in the project repository\footnote{https://github.com/rpiRobotics/ARM-21-02-F-19-Robot-Motion-Program}, 
with Tesseract in-browser simulation built in for trajectory visualization.
User only needs to provide robot kinematics descriptions in Unified Robotics Description Format (URDF) and the target spatial curve, $(\bm{p}^*,\bm{n}^*)$, in a comma-separated values (CSV) file. 
This interface guides the user through the three steps discussed in Section~\ref{sec:approach} and is demonstrated in the supplementary video.

\section{Conclusion and Future Work}

This paper introduces a comprehensive method to generate robot motion programs to track complex spatial curves using two industrial robots.   The objective is to achieve high and uniform path speed while maintaining specified tracking accuracy.
Our approach decomposes the problem into three steps: optimizing robot configuration using continuous robot motion, fitting the robot motion by robot motion primitives, and iteratively updating the waypoints specifying the motion primitives based on the actual tracking error. We have demonstrated the efficacy of the approach on ABB robots in simulation and physical experiments,
\ifresubmissionred{\color{red}\fi
and on FANUC robots in simulation.
\ifresubmissionred}\fi

The experimental setup currently utilizes forward kinematics to calculate the TCP location from the robot joint angles. We are applying the method now to direct measurements of the TCP while the forward kinematics may be imprecise.

\section*{ACKNOWLEDGMENT}
\vspace{-0.5em}
Research was sponsored by the ARM (Advanced Robotics for Manufacturing) Institute through a grant from the Office of the Secretary of Defense and was accomplished under Agreement Number W911NF-17-3-0004. The views and conclusions contained in this document are those of the authors and should not be interpreted as representing the official policies, either expressed or implied, of the Office of the Secretary of Defense or the U.S. Government. The U.S. Government is authorized to reproduce and distribute reprints for Government purposes notwithstanding any copyright notation herein.

\bibliographystyle{IEEEtran}
\bibliography{bib}

\end{document}

%% file: lamacro.tex
\newcommand{\bbull}{\begin{tabbing} \hspace{0.25in}\= \kill}
\newcommand{\ebull}{\end{tabbing}}

\newcommand{\beq}{\begin{equation}}
\newcommand{\eeq}{\end{equation}}
\newcommand{\beqa}{\begin{eqnarray}}
\newcommand{\eeqa}{\end{eqnarray}}
\newcommand{\ba}{\begin{eqnarray*}}
\newcommand{\ea}{\end{eqnarray*}}
\newcommand{\bal}{\begin{align*}}
\newcommand{\eal}{\end{align*}}
\newcommand{\bc}{\begin{center}}
\newcommand{\ec}{\end{center}}
\def\bs{\begin{slide}{}}
\def\es{\end{slide}{}}
\def\bi{\begin{itemize}}
\def\ei{\end{itemize}}
\newcommand{\barr}{\begin{array}}
\newcommand{\earr}{\end{array}}
\newcommand{\lamatrix}{\left[ \begin{array}}
\newcommand{\elamatrix}{\end{array}\right]}
\newcommand{\ma}{\left[ \begin{array}}
\newcommand{\ema}{\end{array}\right]}

\def\bcase{\left\{ \begin{array}{ll}} 
\def\ecase{\end{array} \right.}

\newcommand{\sq}{ ^2 }
\newcommand{\tr}{ ^T }

\newcommand{\abs}[1]{\left| {#1} \right|}

\newcommand{\integral}[2]{\int_{#1}^{#2}}

 1

\newcommand{\comments}[1]{}

\def\clamup{C\kern.1667em\lower-.5ex\hbox{La}\kern.125emMS\ }
\def\clamdown{C\kern-.1667em\lower.5ex\hbox{La}\kern-.125emMS\ }
